\documentclass{article}
\usepackage{spconf,amsmath,epsfig}
\usepackage{textcomp}
\usepackage{amsfonts}
\usepackage{xcolor}

\def\reals{\ensuremath{\mathbb{R}}}

\newcommand{\mat}[1]{\ensuremath{\mathbf{#1}}}

\newcommand{\norm}[1]{\ensuremath{\left\|#1\right\|}}
\newcommand{\cost}[1]{\ensuremath{\ell_{#1}}}

 \newcommand{\refeq}[1]{(\ref{#1})}

\def\dictm{\mat{D}}

\def\datam{\mat{X}}
\def\datav{\mat{x}}

\def\coefm{\mat{A}}
\def\coefv{\mat{a}}

\def\ndims{m}
\def\natoms{p}
\def\nsamples{n}
\def\group{G}

\def\ngroups{{\mathcal{G}}}

\title{Collaborative Sources Identification in Mixed Signals via Hierarchical Sparse Modeling\thanks{Work supported by ONR, NGA, ARO, NSF, and NSSEFF.}}
\name{
Pablo Sprechmann, Ignacio Ramirez, Pablo Cancela, and Guillermo Sapiro\vspace{-.17in}
}
\address{University of Minnesota and Universidad de la Republica, Uruguay}

\begin{document}
\ninept
\maketitle
\vspace{-.1in}
\begin{abstract}
A collaborative framework for detecting the different
sources in mixed signals is presented in this paper. The approach is
based on C-HiLasso, a convex collaborative hierarchical sparse model, and proceeds as follows.
First, we build a structured dictionary for mixed signals by concatenating a set of sub-dictionaries, each one of
them learned to sparsely model one of a set of possible classes. Then, the coding of the mixed signal is
performed by efficiently solving a convex optimization problem that combines
standard sparsity with group and collaborative sparsity. The present
sources are identified by looking at the sub-dictionaries automatically selected in the coding.
The collaborative filtering in C-HiLasso takes advantage of the
temporal/spatial redundancy in the mixed signals, letting collections of samples collaborate in
identifying the classes, while allowing individual samples to have different internal
sparse representations. This collaboration is critical to further stabilize the sparse representation of signals, in particular the class/sub-dictionary selection. The internal sparsity inside the sub-dictionaries, as naturally incorporated by the hierarchical aspects of C-HiLasso, is critical to make the model
consistent with the essence of the sub-dictionaries that have been trained for sparse representation of each individual class.
We present applications from speaker and instrument identification and texture separation.
In the case of audio signals, we use sparse modeling to describe the short-term power spectrum envelopes of harmonic sounds.
The proposed pitch independent method automatically detects the number of sources on a recording.

\end{abstract}

\vspace{-.1in}
\section{Introduction and Motivation} 
\label{sec.intro}

Sparse signal modeling has been shown to lead to numerous state-of-the-art
results in signal processing, in addition to being very attractive at the
theoretical level. The standard model assumes that a signal can be
efficiently represented by a sparse linear combination of atoms from a given
or learned dictionary. The selected atoms form
the {\it active set}, whose cardinality is significantly smaller than the
size of the dictionary and the dimension of the signal.  Adding structural
constraints to this active set has value both at the level of representation
robustness and at the level of signal interpretation; e.g.,
\cite{yuan06,bach09,EM09}. This leads to {\it group}
or {\it structured} sparse coding, the atoms are grouped and a few groups are active at a time. An
alternative way to add structure (and robustness) to the problem is to
consider the simultaneous and collaborative encoding of multiple signals,
requesting that they all share the same active set; e.g.,
\cite{tropp06a,CREK05,CH06}.

In the (linear) source separation problem, an observed signal is
assumed to be a linear superposition (mixture) of several
sources, and the primary task is to estimate from it each of the unmixed sources. If the task is only to identify the active sources,
the problem is called source identification. In this case, since the original sources
do not need to be recovered, the modeling can be done in terms of features extracted from the original
signals in a non-bijective way. We propose to first use traditional sparse modeling tools to learn a dictionary $\dictm_i$ for each one of the $\ngroups $ possible classes.
Concatenating these dictionaries, $\dictm=[\dictm_1| \dictm_2|\ldots|\dictm_{\ngroups}]$, any mixture
signal produced by that ensemble will be accurately represented as a sparse
linear combination of the atoms of this larger dictionary $\dictm$. In this case one expects the resulting
sparsity patterns to have a particular structure, with sub-dictionaries active following the classes present in the mixture. In addition,
the time correlation in audio signals and the spatial correlation in images
suggests that there is an important correlation between neighboring samples that should be exploited.
Consider for example a piece of music, where a few out of many potential instruments are playing simultaneously at different times. For small time windows, we can assume that the same few instruments are playing at all instants (each instant represents a mixture sample), so that the
corresponding same few groups or sub-dictionaries will be active in all
samples. However, we do not expect the sound produced by each instrument to
be the same at each instant, the internal activation per sub-dictionary will be sample dependent.

We propose to use the Collaborative Hierarchical Lasso (C-HiLasso) model that combines the
benefits of structured and collaborative sparse coding in a hierarchical
sparse model, with sparsity both at the group and single
coefficient levels, and where multiple signal samples/instances
collaborate in the recovery of their common active groups. However, for each
signal, the active atoms within the shared active groups are particular to that
signal realization.
The internal sparsity inside the blocks, which is not
present in standard structured/block sparsity models, is critical, since each block corresponds to a sub-dictionary $\dictm_i$ learned to efficiently represent signals of one of the possible classes in a sparse coding fashion. Not considering such in-block sparsity will then be contradictory to the essence of the dictionary model, while not considering sparsity and collaboration at the block level will contradict the
fact that only a few classes are active per instance of the signal, and such classes are shared, e.g., in audio as explained above.
Previously proposed sparsity models, e.g., group or collaborative sparsity, or elastic net \cite{EN}, don't have these characteristics, which are critical for the problem at hand and consistent with the realistic assumptions about the signal. In \cite{JournalHiLasso}
we provide additional details and variations of the proposed model, including
a detailed comparison of C-HiLasso with recent literature, theoretical results regarding recovery guarantees, and an efficient optimization techniques that ensure convergence to the global optimum. The goal of this work is to show how the this framework can be successfully applied to several types of signals by appropriately selecting the features.

In Section~\ref{sec:intro} we briefly describe the CHiLasso model.
In Section~\ref{sec:results:audio} we address the problem of single-channel speaker and instrument identification.
The feature selection is crucial for the success and the efficiency of the model. The proposed method uses the spectral envelope as feature vectors and does not require the estimation of the fundamental frequency of the sources. In Section~\ref{sec:results:textures} we address the problem of texture separation and identification.

\section{Collaborative Hierarchical Sparse Coding}
\label{sec:intro}

We have a set of data samples $\datav_j \in \reals^{\ndims},
j=1,\ldots,\nsamples$, and a dictionary of $\natoms$ atoms in
$\reals^{\ndims}$, assembled as a matrix $\dictm \in
\reals^{\ndims{\times}\natoms}$
. Each sample $\datav_j$ can be written as
$\datav_j=\dictm\coefv_j+\epsilon,\,\coefv_j \in \reals^{\natoms},\,\epsilon
\in \reals^{\ndims}$,
$\norm{\epsilon}_2\ll \norm{\datav_j}_2$. The underlying assumption in
sparse modeling is that the ``optimal'' reconstruction
$\coefv_j$ has only a few nonzero elements.
The convex formulation of this representation, known in the
literature as \emph{Lasso} \cite{tibshirani96},
can be efficiently solved using general purpose or specialized
optimization techniques. A popular variant is the unconstrained
version,
\begin{equation}
\min_\coefv \frac{1}{2}\norm{\datav_j-\dictm\coefv}_2^2 + \lambda\norm{\coefv}_1,
\label{eq:lasso}
\end{equation}
where $\lambda$ is an parameter value, usually found by
cross-validation.

In many situations, one has prior knowledge that certain groups
of atoms are simultaneously selected in the coding. Designing a model that takes this into account naturally leads to a better result.
Suppose that a dictionary of $\natoms$ atoms is divided into $\ngroups$ groups.\footnote{For simplicity we assume that all the groups have the same size. The extension to the general case is straightforward, see \cite{yuan06} for details.}
We refer to the \emph{sub-dictionary} of atoms of $\dictm$ belonging to a group $G$ as
$\dictm_G$, and the corresponding set of linear reconstruction
coefficients as $\coefv_G$.  The \emph{Group Lasso} problem is,
\cite{yuan06},
\begin{equation}
\min_\coefv \frac{1}{2}\norm{\datav_j-\dictm\coefv}_2^2 + \sum_{G=1}^{\ngroups}\norm{\coefv_G}_2.
\label{eq:group-lasso}
\end{equation}
Note that \eqref{eq:group-lasso} reduces to \eqref{eq:lasso} when the groups contain only one atom, and
its effect on the groups of $\coefv$ is a natural generalization of Lasso: it turns on/off coefficients in groups.


In numerous applications, one expects that certain collections of
samples, $\datam=[\datav_1,\ldots,\datav_\nsamples]\in \reals^{\ndims{\times}
  \nsamples}$,  share the same active components from the
dictionary, that is, the indexes of the corresponding nonzero coefficients, $\coefm=[\coefv_1,\ldots,\coefv_\nsamples]\in \reals^{\natoms{\times}
  \nsamples}$, are the same for all the samples in the collection.
Imposing such dependency in the $\cost{1}$ regularized regression
problem gives rise to the so called {\it collaborative} (also called
``multitask'' or ``simultaneous'') sparse coding problem
\cite{wright04}. The model is given by
\begin{equation}
\min_{\coefm} \frac{1}{2}\norm{\datam - \dictm\coefm}_F^2 +
 \lambda \sum_{k=1}^{\natoms}{\norm{\coefv^k}_2},
\label{eq:collaborative-lasso}
\end{equation}
where $\coefv^k\in \reals^{\nsamples}$ is the $k$-th row of $\coefm$, that
is, the vector of the $\nsamples$ different values that the coefficient
associated to the $k$-th atom takes for each sample $j=1,\ldots,\nsamples$.
If we extend this idea to the Group Lasso, we obtain a {\it
  collaborative Group Lasso} (C-GLasso),
\begin{equation}
\min_{\coefm} \frac{1}{2}\norm{\datam- \dictm\coefm}_F^2 +
 \lambda \sum_{\group=1}^{\ngroups}{\norm{\coefm^\group}_F},
\label{eq:collaborative-group-lasso}
\end{equation}
where $\coefm^\group$ is the sub-matrix
formed by all the rows belonging to group $\group$.

As explained in Section~\ref{sec.intro}, in our proposed strategy for performing source separation,
each $\mat{D}_{\group}$ is trained for sparsely representing one of the possible sources in the mixture.
Hence, the sparsity pattern of the coefficients of the mixture signals is hierarchical: sparsity at the group and atom levels.
In this situation the C-GLasso would fail to recover the true sparsity pattern, since it promotes all the atoms in the
sub-dictionary to be selected simultaneously. We also need to consider collaboration at the group level, as in \eqref{eq:collaborative-group-lasso},
but not at the individual atoms level. The only alternative that can handle all these requirements simultaneously is C-HiLasso (see \cite{JournalHiLasso} for details on how to automatically set the regularizer parameters $\lambda_{1,2}$ and also details on the optimization),
\begin{equation}
\min_{\coefm} \frac12 \norm{\datam - \dictm\coefm}_F^2
+   \lambda_2 \sum_{\group=1}^{\ngroups}{\norm{\coefm^\group}_F} +  \lambda_1\sum_{j=1}^{\nsamples} \norm{\coefv_j}_1.
\label{eq:collaborative-hilasso}
\end{equation}
The regularizer in \refeq{eq:collaborative-hilasso} is a combination of the ones used in C-GLasso and Lasso and as such encourages the signals to share the
same groups (classes), while the active set inside each group is signal
dependent.
Note that the last term in  \refeq{eq:collaborative-hilasso} can be replaced by a group sparsifying norm, e.g., if atoms on the sub-dictionary have some correlation.
Previous approaches have only considered particular cases, such as structured coding \cite{bach09}, hierarchy without collaboration
\cite{friedman10a}, or collaboration without hierarchy
\cite{peng,boufounos10}. The comprehensive new model, \cite{JournalHiLasso}, is needed for the important applications presented next.


\section{Source Identification in Audio}
\label{sec:results:audio}

Source identification is a classic problem in audio analysis, see \cite{Schmidt-channelspeech,pep3} and references therein. Here is addressed with the C-HiLasso model.

\subsection{Feature Selection for Speaker Identification}

A challenging aspect when identifying audio sources is to obtain features that
are specific to each source and at the same time invariant to changes in the
fundamental frequency (tone) of the sources.  In the case of speech, sounds
can be divided into two main groups, \emph{voiced} and \emph{unvoiced}
sounds. Of the two, only the former contains information useful for
identifying the speaker.  Since unvoiced sounds have much less energy than
voiced ones, we can easily remove them from the feature extraction process,
so that the identification is performed solely with the voiced sounds. To
describe voiced sounds, we use their short-term power spectrum envelopes (SE) as
feature vectors, which is a common choice in speaker recognition tasks \cite{Speech}.

The SE in human speech varies along time, producing different
patterns for each phoneme. Then, a speaker does not produce a unique
SP for voiced sounds, but a set  that lives in a union of manifolds.
Since such manifolds are well represented by
sparse models, the SE characteristics are well suited for the sparse modeling framework.
For C-HiLasso,
the feature extraction process needs to be linear, and extracting the SE is not a linear operation. To overcome this,
we propose a method inspired on the \emph{Mel Frequency Coefficients (MFCC)} technique
\cite{Speech}, exploiting the harmonic properties of voiced sounds.

\begin{figure}
\begin{center}
\includegraphics[width=0.49\textwidth]{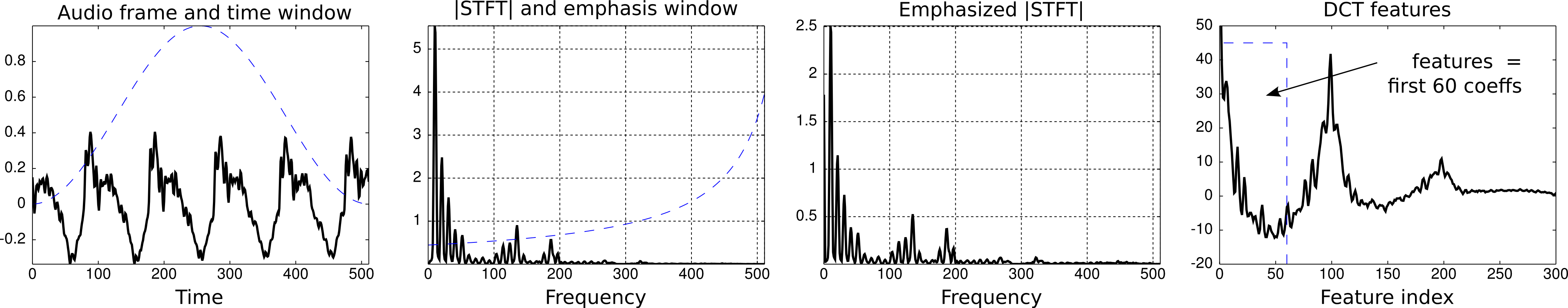} \vspace{-.17in}
\vspace{-2ex}\caption{\footnotesize \it \label{fig:audio-features} Feature extraction for audio signals. From left to right: sample analysis window, its STFT
  magnitude and emphasis curve, emphasized STFT magnitude, DCT and low
  frequency samples used as features (dotted). }\vspace{-.2in}
\end{center}\vspace{-.1in}
\end{figure}

Assume that we observe a signal $y(n)$ that is a linear mixture of $c$
harmonic sources,
$$y(n) = \sum_{i=1}^{c}{ \alpha_i  \sum_{k=1}^{K}{ E_i\left(\frac{2\pi}{f_s} k f_i\right) \cos\left(\frac{2\pi}{f_s} k f_i n + \phi_{ik}\right) }  },$$
where $\alpha_i$ are the mixing coefficients, $E_i$ and $f_i$ are the SE and fundamental frequency of the $i$-th source respectively, $f_s$ is the sampling frequency, and
$\phi_{ik}$ is the phase of the $k$-th harmonic (or partial), out of $K$, of that source. As with MFCC, we start the analysis of an audio window by performing a short-term Fourier transform (STFT) on it. Since phase information is irrelevant for computing the spectral envelope, we only keep the magnitude of the obtained STFT.  In order to amplify the frequency
range of interest, we apply an emphasis window $W(\theta)$ to the STFT magnitude, obtaining
\begin{equation}
|Y(\theta)| = \sum_{i=1}^{c}{ \alpha_i  \sum_{k=1}^{K}{ W(\theta) E_i\left(\frac{2\pi}{f_s} k f_i\right) \delta\left(\theta-\frac{2\pi}{f_s} k f_i \right) }  },
\label{eq:MFCCcoefficients}
\end{equation}
where $\delta$ is the Dirac distribution.\footnote{We only write the positive part of the spectrum.}
Finally, we perform a discrete cosine transform (DCT) on the emphasized STFT magnitude (which is a real function),
obtaining, for each source, the convolution of a low frequency lobe that approximates the spectrum of the emphasized SE, with a
sequence of spikes corresponding to the fundamental frequency and its partials.
Keeping only the low-frequency coefficients of the computed DCT in \eqref{eq:MFCCcoefficients} , we obtain $\sum_{i=1}^{c}{ \alpha_i DCT\left\{ W(\theta)E_i(\theta)\right\} }$, which is a linear combination of the spectrums of the emphasized SE of the present sources.
We then obtained a linear relation between the sources and their spectral envelopes. The feature extraction process is summarized in Figure~\ref{fig:audio-features}.

The audio files were re-sampled at $f_s=16kHz$. The features
were taken based on the STFT with a frame length of 512 samples and an
overlapping of 75\% using a Hanning window. The emphasis in frequency is
$E(f) = 1+\alpha f$, $\alpha = 2/f_s$. After the DCT, the lowest 60 coefficients form the features.

\begin{figure*}
\begin{center}
{\footnotesize
\begin{tabular}[b]{|c|ccc|}\hline
   & 5-NN & 25-NN& \cite{Sprechmann_ICASSP} \\ \hline
F1 & 17.8 & 19.9 & 17.7 \\
F2 & 19.0 & 17.6 & 20.5 \\
M1 &  9.5 &  9.0 & 15.1 \\
M2 & 16.9 & 18.2 & 13.9 \\
M3 & 24.3 & 28.3 & 17.6 \\
{\it AVG}& {\it 17.5} & {\it 18.6} & {\it 17.0} \\ \hline
 \end{tabular}%
}
\hspace{.1in}\includegraphics[height=0.1\textheight]{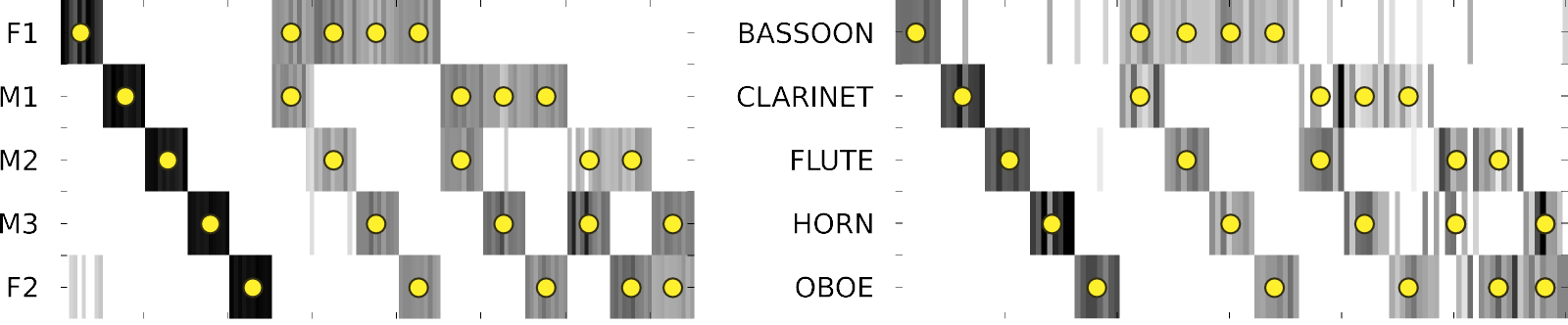}%
\caption{\footnotesize \it \label{fig:audio-results} Results for audio
  source identification.  The leftmost table shows single source detection
  results for a simple k-NN classifier (with $k=5$ and $25$) and the sparse
  model classifier presented in \cite{Sprechmann_ICASSP}.  The center and right figures
  show identification results obtained with C-HiLasso when the sources are
  speakers (left) and wind instruments (right). Each column of the graph
  corresponds to the sources identified for a specific time frame, with the true ones marked by yellow dots. The
  vertical axis indicates the estimated activity of the different sources,
  where darker colors indicate higher energy. In the speaker identification case,
  we have 10 frames (15 seconds of audio) for each possible combination and
  in the instrument case, 8 frames (3 seconds).
  For speakers we used $(\lambda_1,\lambda_{2_0})=(0.8,0.008)$ and for the instruments $(\lambda_1,\lambda_{2_0})=(0.8,0.015)$. We observe how the number and type of classes are correctly identified.}\vspace{-.3in}
\end{center}
\end{figure*}

\subsection{Speaker Identification via C-HiLasso}
\label{sec.speak.ident}

The data for this case consists of six minutes long recordings of five different German radio
speakers, two female and three male.\footnote{The dataset is available from the authors upon request.} One quarter of the samples are used for training and the rest for testing.

First we want to ensure that the proposed features and the sparse
modeling framework are well suited for this application.  We analyzed the dataset using two very simple classifiers: $k$ nearest
neighbors, and the classifier proposed in \cite{Sprechmann_ICASSP}. In the
latter case, following standard dictionary learning techniques, a dictionary
$\mat{D}_i$ is learned for each class using the corresponding training
samples.  Each normalized testing sample, $\mat{x}_j$, is then assigned to the class
for which the risk
$R(\mat{D}_i,\mat{x}_j)= \min_{\mat{a}_j} \norm{\mat{x}_j -
 \mat{D}_i \mat{a}_j} + \lambda \norm{\mat{a}_j}$
is minimized ($\lambda$ is
learned via cross validation). The error rate obtained for each
speaker by each method is shown in
Figure~\ref{fig:audio-results}(left). Although a per-sample error rate of $17.0\%$ is not small, each
sample corresponds to a time window of 32ms and we can safely assume that
the same speaker will be active during several consecutive samples. Thus a
simple voting scheme on top of any of these classifiers would reduce the
error significantly. %
The collaboration, naturally included in C-HiLasso, is crucial for
the identification, also when mixtures are present as detailed next.

In the analysis above we assumed the strong hypothesis that only one speaker
is active at a time. We now relax this and
test the performance of C-HiLasso in identifying speakers in mixture
signals. Clearly, $k$ nearest neighbors can't be used in this case.  For
each speaker, a sub-dictionary of $90$ atoms was learned from the training
dataset (we observed that the exact dictionary size is not critical to the results of the algorithm).
We extracted $10$ non-overlapping frames of 15 seconds each,
and encoded them using C-HiLasso.  The experiment was repeated for all
possible combinations of two speakers, and all the speakers talking
alone. In order to quantify the performance we measured the Hamming distance
between the detected active sources and the ground truth.
We compared the results only against the Lasso algorithm. This is the canonical experiment since the dictionaries where trained
to sparsely represent the data (C-GLasso assumes that all the atoms in the sub-dictionary are active simultaneously).
C-HiLasso obtained a Hamming distance of $0.053$, showing a very good
capability of automatically detecting the active sources (speakers) on each
frame without having the number of active sources as prior knowledge.
Lasso gives a Hamming distance of $3.33$. The Hamming distance
when there is only one speaker in the mixture signal is $0.08$ and $0.04$ when there are two of them.
Again here the Lasso performs worse giving $4$ and $3$ respectively.

In Figure~\ref{fig:audio-results} we show the results obtained for
each frame. One could think of adding robustness to this method by evaluating overlapping time frames and doing time
regularization.

\subsection{Instrument Identification}
\label{sec.inst.ident}

Unlike the case of the human voice where the fundamental frequency can vary
over a small range of values, in musical instruments it can vary
considerably from one instrument to the other. For example in the experiment
bellow, the fundamental frequencies vary from 80Hz (bassoon) to 1600 Hz (flute). The above proposed features
represent a good description of the spectral envelope for low fundamental
frequency sources. When considering sounds with high fundamental frequency,
the descriptor represents a mixture of information of the envelope and
fundamental frequency together.  This happens because a non-adaptive linear
operation can't separate them for a very wide range of fundamental
frequencies. At first glance this may appear as a drawback, but in fact it
becomes an advantage as it includes some information from the fundamental
frequency into the descriptor, still keeping reasonable dictionary
size.

We used the \emph{Development Set for MIREX 2007 MultiF0 Estimation Tracking
  Task},\footnote{\texttt{http://www.music-ir.org/mirex/wiki/}.}  which
consists of a 52 second long musical piece played by five different wind
instruments (bassoon, clarinet, flute, horn and oboe), and a set of tracks
where each instrument plays individually. We used the first half of these
audio tracks for training, and the rest for testing.  In some passages of
this piece, the instruments are arranged harmonically (forming chords),
meaning that the notes one plays are partials of the fundamental note played
by others. Thus, in these passages, the partials of the intervening
instruments superimpose.

The experiment for this case is analogous to the one with speakers, with the
testing tracks divided in frames of 3 seconds each. The results are shown in
Figure~\ref{fig:audio-results}. The average Hamming distance between the
identified sources and the ground truth for C-HiLasso and Lasso was  respectively $0.16$ and $2.46$ when only one source was
active, and $0.18$ and $2.76$ for all combinations of two sources, for a total average
of $0.17$ and $2.56$. Once again, this demonstrates the power of C-HiLasso in
collaboratively identifying the correct instruments (classes or
sub-dictionaries). The hierarchical component is critical since, while all
the signals share the active instruments (and the sub-dictionaries), each
time frame is different, meaning they are represented using different atoms
of the detected sub-dictionaries.

\section{Source Separation in Texture Images}
\label{sec:results:textures}

\begin{figure}
\begin{center}
\includegraphics[width=0.4\textwidth]{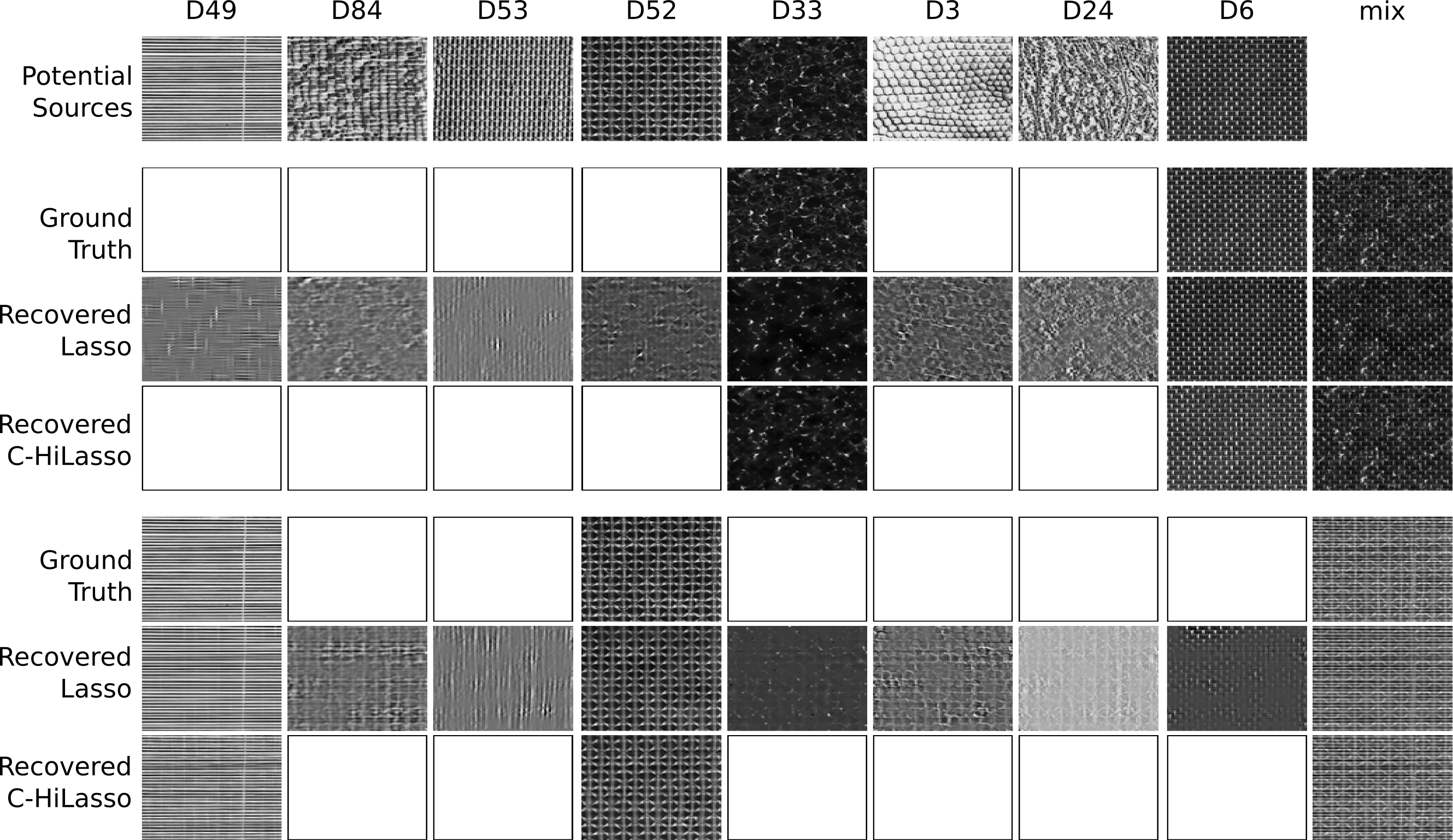}
\end{center}
\vspace{-2ex}
\caption{%
\label{fig:texture-separation}%
\footnotesize%
\it Texture separation results. We show only two sample combinations out of the
28 possible combinations of 2 active sources out of 8 (inactive ones are
shown in white). Standard Lasso,
which is not designed for this task, recovers residual sources which were not
present in the original mixtures. The quality of the recovered active
textures, although not evident from the thumbnails above, is also 2dB better
in both examples, and 1dB on average for all combinations, compared to what
we can obtain with Lasso.}\vspace{-.2in}
\end{figure}

\vspace{-.1in}
Using sparse modeling for addressing the source separation problem in images has been addressed in  \cite{EladTexture,bobin}. The methods are designed for non-collaborative separation of mixtures of two given classes. In this section, we explore the capabilities of C-HiLasso for source separation in
images which are mixtures of a few out of several possible textures drawn
from the Brodatz dataset, Figure~\ref{fig:texture-separation}.\footnote{\texttt{http://www.ux.uis.no/\~{
    }tranden/brodatz.html}} The columns of $\mat{X}$ contain the pixel values of
all possible square windows of $10{\times}10$ pixels in the mixture image as $\ndims=10^2$ dimensional vectors. The sub-dictionaries
$\dictm_G$ for each texture source were obtained off-line from training
samples taken from the left halves of the texture images, while the samples
used in the tests were taken from the right halves.  Clearly, if the image
is a mixture of a number of source texture images, then
every sample will also be a mixture of the same corresponding classes in the
source images, and the hypothesis of C-HiLasso will hold.  The experiment
was repeated for all possible $28$ combinations of $2$ out of $8$ possible
source textures. In terms of detected groups, the C-HiLasso
achieves near perfect performance, with an average Hamming error between the
true and estimated active sets of $0.14$. Lasso is clearly not designed for
this task, yielding an average Hamming error of $2.8$.
The best average PSNR (APSNR) obtained with C-HiLasso for all
combinations was $23.7dB$, which is $2dB$ larger than the $21.7dB$ obtained
with Lasso.  The C-GLasso obtains a Hamming error of $0.62$ (three times larger than C-HiLasso),
and gives a significantly lower APSNR of $19.8dB$, clearly showing that the model is not good for representing the data.

We conclude that C-HiLasso is efficient for collaboratively identifying sources in a set of
mixture signals. The framework is capable of
identifying sources in audio and identifying and recovering mixed
sources in images, always detecting the number of
sources present in the mixture.

\footnotesize
\bibliographystyle{IEEEbib}
\footnotesize
\bibliography{cvxopt}

\end{document}